\documentclass{article} 
\usepackage[final]{colm2026_conference}

\usepackage{microtype}
\usepackage{hyperref}
\usepackage{natbib}
\usepackage{url}
\usepackage{booktabs}
\usepackage{graphicx}
\usepackage{amsmath}
\usepackage{placeins}
\usepackage{comment}
\usepackage{capt-of}
\usepackage[T1]{fontenc}
\usepackage{lmodern}
\usepackage{microtype}
\usepackage{xcolor}
\usepackage{pifont}      
\usepackage{ifthen}
\usepackage[most]{tcolorbox}
\usepackage{subcaption}

\usepackage{lineno}
\usepackage{hyperref}

\definecolor{darkblue}{rgb}{0, 0, 0.5}
\hypersetup{colorlinks=true, citecolor=darkblue, linkcolor=darkblue, urlcolor=darkblue}

\title{Decodability is Not Causality: Dissociating Probe Readouts from Behavioral Drivers via SAE Decomposition}

\author{
\textbf{Devesh Tiwari\textsuperscript{1,}\kern0.08em\thanks{These authors contributed equally to this work.},
Camille Davis\textsuperscript{2,}\kern0.08em\footnotemark[1],
Shivank Sinha\textsuperscript{3}}
\\
\textbf{Talia Weaver\textsuperscript{4},
Aditya Shah\textsuperscript{5},
Maheep Chaudhary\textsuperscript{6,}\kern0.08em\thanks{Senior author.}}
\\
\textsuperscript{1}WW-P High School South \quad
\textsuperscript{2}Phillips Academy Andover \quad
\textsuperscript{3}Dublin High School \\
\textsuperscript{4}Carlmont High School \quad
\textsuperscript{5}Google \quad
\textsuperscript{6}Independent
\\
\texttt{\{camillehdavis\}@Gmail.com},
\texttt{\{deveshtiwari2705\}@Gmail.com}
}

\begin{document}

\ifcolmsubmission
\linenumbers
\fi

\maketitle

\begin{abstract}
Linear probes can decode safety-relevant concepts such as truthfulness from language-model activations, but probe accuracy may show only decodability, not that the features the probe weights causally drive model behavior. We demonstrate that this gap cannot be closed from the geometry of probe weights alone: the features geometrically aligned with probe direction need not be the ones the model uses, so causal relevance requires intervention. We introduce a feature-level diagnostic that decomposes a deployed \textit{True}/\textit{False} probe into sparse-autoencoder (SAE) features, ranks those features by both probe alignment and by gradient sensitivity of the model's behavior, and ablates the resulting shared, probe-only, and random feature sets under a coherence gate. On the truth probe of \citet{burger2024truth} (TTPD), applied in the instructed truth/deception setting of \citet{long2025truthfulrepresentationsflipdeceptive} for Gemma-2-9B-Instruct, the two rankings overlap only weakly (about \(12\%\), Spearman \(\rho = 0.10\)), and ablation dissociates them sharply: features the probe shares with the model flip the output far more (up to \(27\%\)) than equally sized probe-only (\(6\%\)) or random (\(1\%\)) features at full coherence, while probe-only features instead perturb the probe's own readout. The dissociation holds across five seeds and a held-out split, and an activation-aware selection of features flips behavior nearly three times as often as the probe's geometric top features (\(17.6\%\) vs. \(6.1\%\)). In this setting, therefore, the geometric projection of a probe's weight vector alone does not identify the features the model causally uses; however, combining probe information with feature activation statistics recovers substantially more behaviorally causal features, and coherence-gated SAE intervention is needed to separate them from probe readouts. We make the implementation of our methods accessible at \href{https://github.com/cam1lled/causal-validation-sae}{this URL}.


\end{abstract}

\section{Introduction}
Linear probes trained on model activations are widely used to identify and predict representations associated with behavioral states \citep{alain2018intermediate,belinkov2022probing}.
However, information that is decodable from a model's activations is not necessarily causally responsible for the model's behavior. Consequently, high probe accuracy alone does not establish that the representations identified by the probe necessarily play a causal role in the model's decisions \citep{hewitt2019designing,belinkov2022probing,occhipinti2026probing}. Distinguishing between features that are merely predictive and those that are causally relevant remains an open challenge.


We investigate whether probe-identified features are causally involved in the model behaviors associated with the representations they decode. Using the truthful/deceptive factual-verification setting introduced by \cite{long2025truthfulrepresentationsflipdeceptive}, we decompose model activations via a sparse autoencoder (SAE) and independently rank features according to both probe alignment and gradient sensitivity of the model's output. We then perform direct interventions on these feature sets while tracking the output coherence.  

We find that the features the probe weights most and the features the model is sensitive to overlap only weakly, and that this difference is behavioral. Ablating the features the probe shares with the model flips the model's output substantially more than ablating equally sized sets of probe-only or random features, while output coherence is preserved throughout. Probe-only features instead perturb the probe's own readout without changing behavior. These results suggest that many probe-weighted features function primarily as diagnostic readouts rather than causal drivers of behavior, and that the geometry of a probe's weights alone does not identify the features the model causally uses.

Our work makes three main contributions. First, we introduce a framework for comparing probe-aligned and gradient-sensitive SAE features through direct behavioral interventions. Second, we quantify the extent to which probe-ranked features produce coherent behavioral effects, showing that predictive importance and behavioral importance can diverge substantially. Third, we identify a subset of shared probe-model features that consistently influence both probe predictions and model behavior, providing stronger evidence of behavioral relevance than probe-only features.
\begin{figure}[tb]
    \centering
    \includegraphics[width=0.8\linewidth]{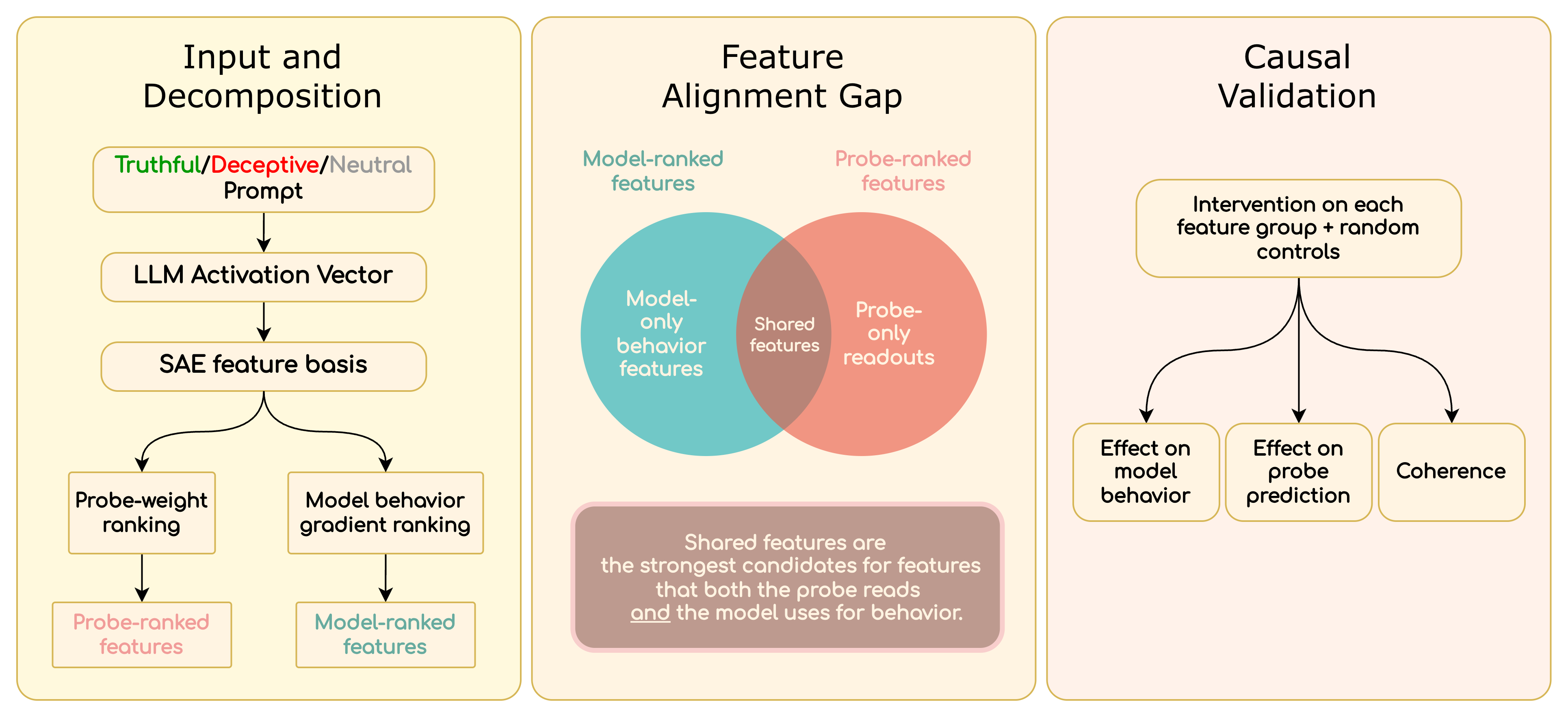}
    \caption{We decompose final pre-generation residual activations into an SAE feature basis, then rank the same SAE features by probe-weight alignment, indicating features the probe reads, and by gradient sensitivity of the model's \textit{True}/\textit{False} margin, indicating features whose perturbation most affects the model's output. We then intervene on each feature group, together with random controls, and measure effects on model behavior, probe prediction, and output coherence.}
    \label{fig:maindig}
\end{figure}

\section{Related Work}

\paragraph{Probing, truth representations, and behavioral use.}
Linear probes are widely used to identify information encoded in neural representations \citep{alain2018intermediate,belinkov2022probing}, but
decodability alone does not establish that the decoded information is used by the model \citep{hewitt2019designing}. Amnesic probing made this distinction explicit by removing probe-identified information and measuring downstream behavior, finding that conventional probing performance need not correlate
with task importance \citep{elazar2021amnesic}. Subsequent work has emphasized that such interventions must be both complete with respect to the targeted property and selective with respect to unrelated information \citep{canby2025reliable}.

For factual truth, prior work has identified approximately linear truth representations and tested their behavioral relevance through activation interventions \citep{marks2024geometry}. Bürger et al. identify a polarity-aware truth subspace and introduce TTPD
\citep{burger2024truth}; Long et al. apply this framework under truthful, neutral, and deceptive instructions and show that truth-related representations change under instructed deception
\citep{long2025truthfulrepresentationsflipdeceptive}. Related probes have been used to detect or steer chain-of-thought unfaithfulness \citep{occhipinti2026probing}, while other studies show that highly accurate probes can instead exploit task-format confounds \citep{sahoo2026linearprobes}. Our work begins from a probe direction that is itself behaviorally causal and asks a finer question: whether the SAE features most geometrically aligned with that direction are the features carrying its behavioral effect.

\paragraph{Sparse autoencoders and causal feature evaluation.}
SAEs decompose model activations into sparse latent features
\citep{cunningham2023sparse,lieberum2024gemmascopeopensparse}, but
reconstruction quality and feature interpretability do not by themselves guarantee causal or task-level utility. Task-grounded evaluations find that unsupervised SAE dictionaries can provide weaker behavioral control than supervised feature dictionaries \citep{makelov2025principled}, struggle to disentangle independently manipulable factual attributes \citep{chaudhary2024evaluatingsaes}, and do not consistently improve over non-SAE probing baselines \citep{kantamneni2025saesuseful}. SAEBench similarly finds that improvements on unsupervised proxy metrics do not reliably transfer
to downstream interpretability tasks \citep{karvonen2025saebench}. These results motivate evaluating SAE features through task-specific interventions and verifying that the studied probe signal is preserved by the SAE reconstruction.

Prior work has used SAE ablation, patching, and steering to identify causally relevant features and circuits \citep{kissane2024interpreting,marks2025sparse}. Most directly, Ma et al. show that contrastively selected SAE reasoning features frequently reflect lexical correlates rather than reasoning
computations \citep{ma2026reasoningfeatures}.

\section{Methodology}

\subsection{Overview}
Our framework proceeds in the following stages: probe replication, SAE feature attribution, gradient-based sensitivity ranking, causal interventions (ablation and learned sparse mask), and overlap and decomposition. 

We track output coherence throughout all interventions to ensure observed effects are interpretable (Section~\ref{sec:coherence}).

We also run auxiliary diagnostics on an additional deception probe to test whether the same SAE-based probe-attribution workflow is suitable for a hardcoded-code deception probe.

\subsection{Stage 1: Probe Replication}
We replicate the target probe using the original authors' procedure, confirming it matches their reported performance (Appendix~\ref{long2025truthfulrepresentationsflipdeceptive_reproduce}).
 
\subsection{Stage 2: SAE Feature Attribution}

We project the probe's weight vector into the decoder basis of a pre-trained SAE to identify which SAE features the probe reads from.

Let the probe $P$ have weight vector $w \in \mathbb{R}^d$, trained on residual stream activations $h^{(\ell)} \in \mathbb{R}^d$ at layer $\ell$. Let the SAE have decoder matrix $D \in \mathbb{R}^{d \times k}$, with columns $d_i$ denoting learned feature directions. Since the probe's output can be approximated as $w^\top h \approx \sum_i f_i \cdot (w^\top d_i)$, where $f_i$ is the activation of SAE feature $i$, the structural contribution of each feature to the probe is:
\begin{equation}
c_i = w^\top d_i
\label{eq:probe-align}
\end{equation}
Features with large $|c_i|$ are directions to which the probe is linearly sensitive within the SAE reconstruction. We rank features by the activation-free score $|c_i|$ as the probe-alignment ranking.

\subsection{Stage 3: Gradient-Based Sensitivity Ranking}
To estimate which features the model's output is most sensitive to, we rank SAE features by a gradient-based attribution of the model's behavior with respect to feature activations. Let $A$ be the attribution objective, defined as the gradient of the behavior logit (or target-token margin) with respect to each feature's activation $f_i^\ell$ at layer $\ell$, averaged over the dataset:
\begin{equation}
A_i = \mathbb{E}_{x \in \mathcal{D}} \left[ \left| \frac{\partial \mathcal{M}(x)}{\partial f_i^\ell} \right| \right],
\end{equation}
where $\mathcal{M}(x) = \operatorname{logit}(\mathrm{True}) - \operatorname{logit}(\mathrm{False})$ is differentiated with respect to the SAE feature activations $f_i^\ell$ at the final-token position of layer $\ell$, averaged over the dataset. We retain the top-ranked features as the gradient-sensitive feature set $\mathcal{F}_{\text{model}}$.

\subsection{Stage 4: Causal Interventions}
Having ranked features from both the probe's perspective and by gradient sensitivity, we validate these attributions through direct intervention. Under each intervention, we measure the behavior flip rate, defined as the fraction of examples where the model's \textit{True}/\textit{False} output changes relative to the unintervened baseline, and the probe-readout shift, defined as the relative change in the probe margin $w^\top h$.

This dual measurement separates a feature's effect on the probe's readout from its effect on the model's behavior, the dissociation at the center of our analysis (Figure~\ref{fig:maindig}).

\subsubsection{Feature Ablation}
For a selected feature set $\mathcal{S}$, we ablate those features by subtracting their reconstructed contributions from the residual stream at layer $\ell$:
\begin{equation}
\tilde{h}^{(\ell)} = h^{(\ell)} - \sum_{i \in \mathcal{S}} f_i \, d_i
\label{eq:ablation}
\end{equation}
and allow the model's forward pass to continue from $\tilde{h}^{(\ell)}$. We apply this procedure to probe-attributed features ($\mathcal{F}_{\text{probe}}$), gradient-sensitive features ($\mathcal{F}_{\text{model}}$), and random control features of matched cardinality.

\subsubsection{Learned Sparse Mask}
We learn a sparse mask over SAE features that selects those contributing most to the probe margin, combining each feature's decoder alignment $c_i$ with its activation. A learnable weight $a_i \in [0,1]$ per feature is optimized with an $\ell_1$ penalty to reconstruct the probe margin from the weighted feature contributions. After training, 30 features exceed $a_i > 0.5$; we ablate the top 16 to match the size of the other feature sets in Table~\ref{tab:functional_mask}. This yields an activation-aware feature set $\mathcal{F}_{\text{mask}}$, in contrast to the activation-free geometric ranking $|c_i|$. Full training details are in Appendix~\ref{sec:mask-details}.


\subsection{Stage 5: Overlap and Decomposition}
We compare the probe's feature ranking against the model's to characterize the extent to which the two rely on shared features.

We compute set overlap between the top-$N$ probe-attributed and gradient-sensitive features as a function of $N$, and report the Spearman rank correlation as a summary statistic.
Using this analysis, we partition the probe's features into three groups: \textbf{Shared}, the intersection of the top-$K$ probe-aligned and top-$K$ gradient-sensitive features, whose size $m$ sets the matched size for the other two groups; \textbf{Probe-only}, comprising the highest probe-ranked features that are not among the model's top gradient-ranked features, size-matched to the shared set; and \textbf{Random}, comprising a size-matched random sample from the remaining features.

We ablate each group at $K \in \{32, 64, 128, 256, 512, 1024\}$ top features from each ranking (yielding shared-set sizes of 5 to 126), subtracting reconstructed feature contributions as in Eq.~\ref{eq:ablation}, and compare the resulting behavior flip rate, probe-margin shift, and coherence. This decomposition isolates the causal contribution of features the probe shares with the model from those unique to the probe.

\subsection{Coherence Gating}
\label{sec:coherence}
Interventions that disrupt the model's ability to produce well-formed output are not informative about the causal role of specific features. We therefore track output coherence under all interventions. We define coherence as the percentage of the responses in which the model responds with either \textit{True}/\textit{False}. An intervention's behavioral effect is only interpreted when coherence is maintained, ensuring that observed behavioral effects are attributable to intervention rather than to general degradation of the model's output. Beyond output validity, we verify three further conditions. First, the SAE faithfully represents the activations it decomposes: its reconstruction attains mean cosine similarity $0.89$ with the original layer-20 activations (normalized reconstruction MSE $0.21$), so interventions act in a basis that captures most of the activation signal rather than in reconstruction noise. Second, the interventions are small targeted perturbations rather than gross corruptions: ablating the shared set removes only $5.7\%$ of the residual-stream norm on average, yet produces the behavioral effects reported in Section~\ref{dissociation}. Third, the interventions leave the model's next-token distribution largely intact: the mean KL divergence between the baseline and post-ablation distributions is $0.11$ nats for the shared set ($0.05$ for the geometric top-16, near zero for random controls; $n=600$), indicating that even the behaviorally effective shared-set ablation is a localized shift of the output distribution rather than a wholesale disruption. 

\section{Experimental Setup}

\subsection{Probe Selection}

\subsubsection{Task and Concept}
We apply our framework to the instructed deception setting of \citet{long2025truthfulrepresentationsflipdeceptive}. A language model is presented with factual statements and prompted under one of three instruction conditions: Truthful, Deceptive, or Neutral. Under the deceptive condition, the model is instructed to respond incorrectly. \citet{long2025truthfulrepresentationsflipdeceptive} train linear probes on the model's residual stream activations to predict the instructed output under each condition, and find that the output is linearly decodable across all three conditions, with accuracy peaking around layer 14 for LLaMA-3.1-8B-Instruct and layer 21 for Gemma-2-9B-Instruct \citep{long2025truthfulrepresentationsflipdeceptive}. 

Appendix~\ref{sec:obfatlasprobe} reports an auxiliary diagnostic on a reward-hacking code probe. The probe is trained on hardcoded versus correct code completions from an MBPP reward-hacking environment, where the behavior of interest is generated code that passes visible tests while failing hidden tests.

\subsubsection{Probe Relevance}
We select the TTPD probe as our primary case study for several reasons. First, the task produces a binary output (\textit{True}/\textit{False}), providing a clean behavioral signal for measuring our intervention effects. The probe also targets a well-defined form of deception: truthful versus deceptive instruction-following in factual verification. The setup is also technically suitable for our method because it uses fixed pre-generation residual-stream activations and has matching GemmaScope SAEs.


\subsubsection{Model, Dataset and SAE}
We use \texttt{Gemma-2-9B-Instruct} \citep{gemmateam2024gemma2} in \texttt{bfloat16} precision. We use the factual-statement datasets from \citet{burger2024truth} (as used by \citet{long2025truthfulrepresentationsflipdeceptive}). We use six affirmative factual-statement sets (\texttt{cities, sp\_en\_trans, inventors, animal\_class, element\_symb, facts}) and their negated counterparts, amounting to twelve sets in total, all under the deceptive instruction condition. Activations are extracted at the final prompt token of layer~20, adjacent to the layer-21 accuracy peak reported by \citet{long2025truthfulrepresentationsflipdeceptive} and the layer with a matching GemmaScope SAE (SAE for Gemma-2-9B-Instruct with a width of 16,384 features from \citet{lieberum2024gemmascopeopensparse}). The behavioral evaluation set comprises 1{,}600 statements; the held-out analysis (Section~\ref{dissociation}) derives the probe direction and feature rankings on one half and measures ablation on the other.

\subsubsection{Behavioral Signal and Coherence}
We define the behavioral signal as
\[
M(x) = \operatorname{logit}(\mathrm{True}) - \operatorname{logit}(\mathrm{False})
\]
at the final token of the templated prompt, the position from which \textit{True}/\textit{False} is generated.

A behavior flip occurs when the sign of this margin under intervention differs from the unintervened baseline. Coherence is the fraction of outputs that remain a valid \textit{True}/\textit{False} token (Section~\ref{sec:coherence}).

\subsubsection{Intervention Configurations}
For feature ablation, we evaluate the following feature sets at $K \in \{32, 64, 128, 256, 512, 1024\}$: probe-attributed ($\mathcal{F}_{\text{probe}}$), gradient-sensitive ($\mathcal{F}_{\text{model}}$), the learned sparse mask ($\mathcal{F}_{\text{mask}}$), shared features, probe-only features, and size-matched random controls.
\section{Results}


\subsection{Probe-aligned and gradient-sensitive features exhibit minimal overlap}
Two rankings, (1) geometric alignment with the probe direction $|w \cdot d|$ and (2) gradient sensitivity of the \textit{True}/\textit{False} margin, identify largely different features. The top 16, 50, 100, and 200 probe features share only 2, 7, 12, and 22 features with the gradient-sensitivity ranking, against chance values of 0.02, 0.15, 0.61, and 2.44 (Spearman $\rho = 0.10$). The shared fraction holds close to 11 to 14 percent across scales, one to two orders of magnitude above chance, but a small minority of either set (Table~\ref{tab:probe-model-overlap}).

\begin{table}[tb]
\begin{center}
\begin{tabular}{lccc}
\toprule
\multicolumn{1}{c}{\bf Top-\(k\) features} &
\multicolumn{1}{c}{\bf Shared features} &
\multicolumn{1}{c}{\bf Shared fraction} &
\multicolumn{1}{c}{\bf Chance overlap} \\
\midrule
16  & 2  & \(12.5\%\) & \(0.02\) \\
50  & 7  & \(14.0\%\) & \(0.15\) \\
100 & 12 & \(12.0\%\) & \(0.61\) \\
200 & 22 & \(11.0\%\) & \(2.44\) \\
\bottomrule
\end{tabular}
\end{center}
\caption{Overlap between geometric probe ranking and model-gradient ranking. Geometric alignment is measured by \(\lvert w \cdot d\rvert\), while model ranking uses gradient sensitivity of the \textit{True}/\textit{False} margin. The model ranking correlates weakly with mean absolute SAE feature activation, with Pearson \(r = 0.07\).}
\label{tab:probe-model-overlap}
\end{table}

Appendix~\ref{overlapfig} plots the full overlap curve.

\subsection{Probe alignment alone does not predict behavioral effect.}

Under ablation at the deceptive prompt, the three feature sets dissociate sharply. Shared features (top probe $\cap$ top model) flip the \textit{True}/\textit{False} output on 13 to 27 percent of statements (0.130 at 16 features, 0.269 at 54);  an equal number of probe-only features (probe-ranked, with the shared model-ranked features removed) flip 6 percent (0.061); random features flip 1 percent (0.010). Exemplars of output changes from these interventions are shown in Appendix~\ref{empiricalexamples}. Coherence remains 1.00 throughout, demonstrating answer flips rather than ablation-induced degradation. The probe's own projection shifts most under probe-only ablation (relative shift 0.24 against 0.05 for shared), as expected given that probe-only features are selected for alignment with the probe direction.

The shared-ablation effect is non-monotonic, rising with set size to a peak near $96$ features before declining as the lowest-ranked shared features shift the margin in the opposing direction and partially cancel the flip. The decline reflects sign-incoherence among these low-ranked features rather than output degradation: the mean absolute margin remains between $1.7$ and $2.8$ throughout, indicating the model continues to emit well-formed, confident \textit{True}/\textit{False} outputs at every ablation count (Appendix~\ref{margin-dip}). The set-level numbers we report ($\le 54$ shared features) lie within the rising regime.

\begin{figure}[tb]
    \centering
    \begin{subfigure}[tb]{0.32\linewidth}
        \centering
        \includegraphics[width=\linewidth]{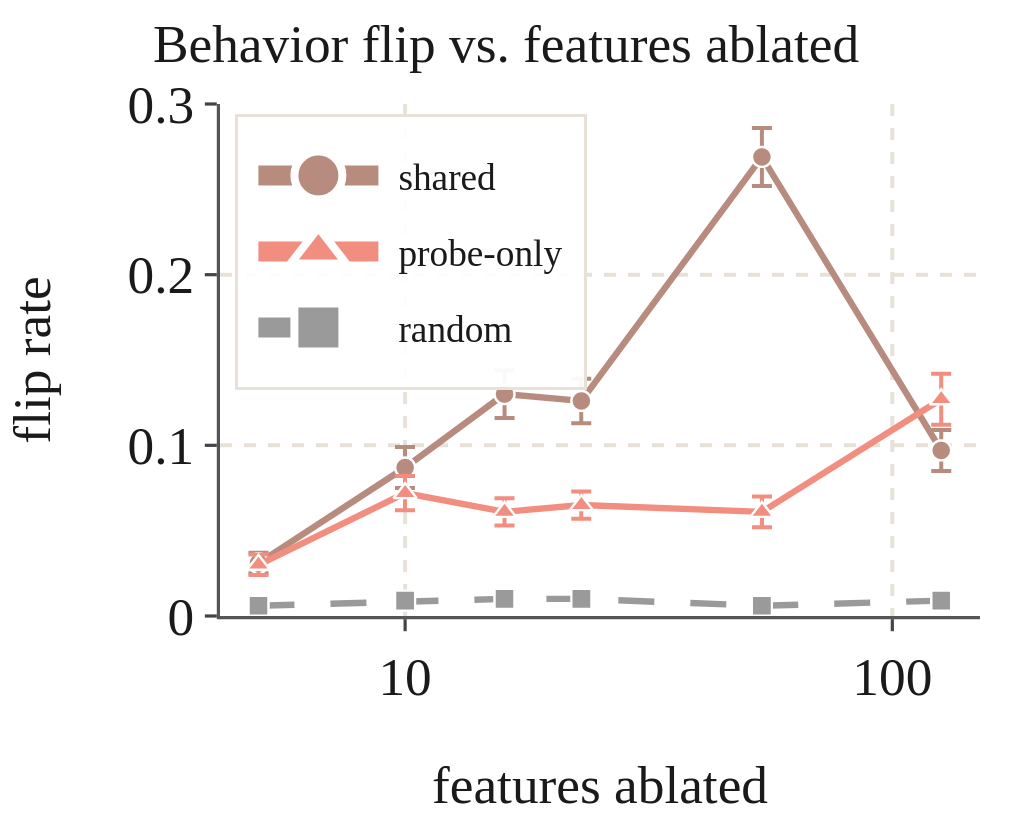}
        \caption{Behavior flip rate. Ablating shared features (top probe $\cap$ top gradient) flips the \textit{True}/\textit{False} output far more than probe-only or random sets of equal size.}
        \label{fig:ablation-flip}
    \end{subfigure}
    \hfill
    \begin{subfigure}[tb]{0.32\linewidth}
        \centering
        \includegraphics[width=\linewidth]{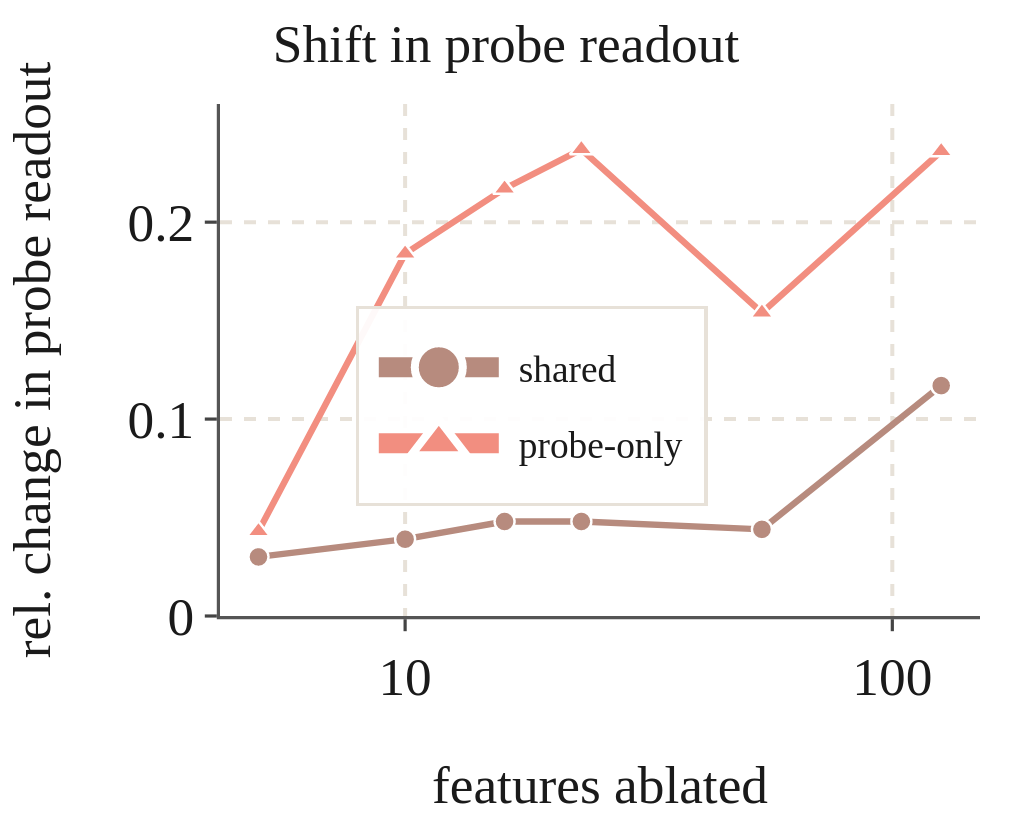}
        \caption{Relative shift in the probe's readout, \(\frac{|\Delta \mathrm{projection}|}{\mathrm{baseline}}\). Probe-only ablation perturbs the readout most, consistent with these features being selected for alignment with the probe direction.}
        \label{fig:ablation-readout}
    \end{subfigure}
    \hfill
    \begin{subfigure}[tb]{0.32\linewidth}
        \centering
        \includegraphics[width=\linewidth]{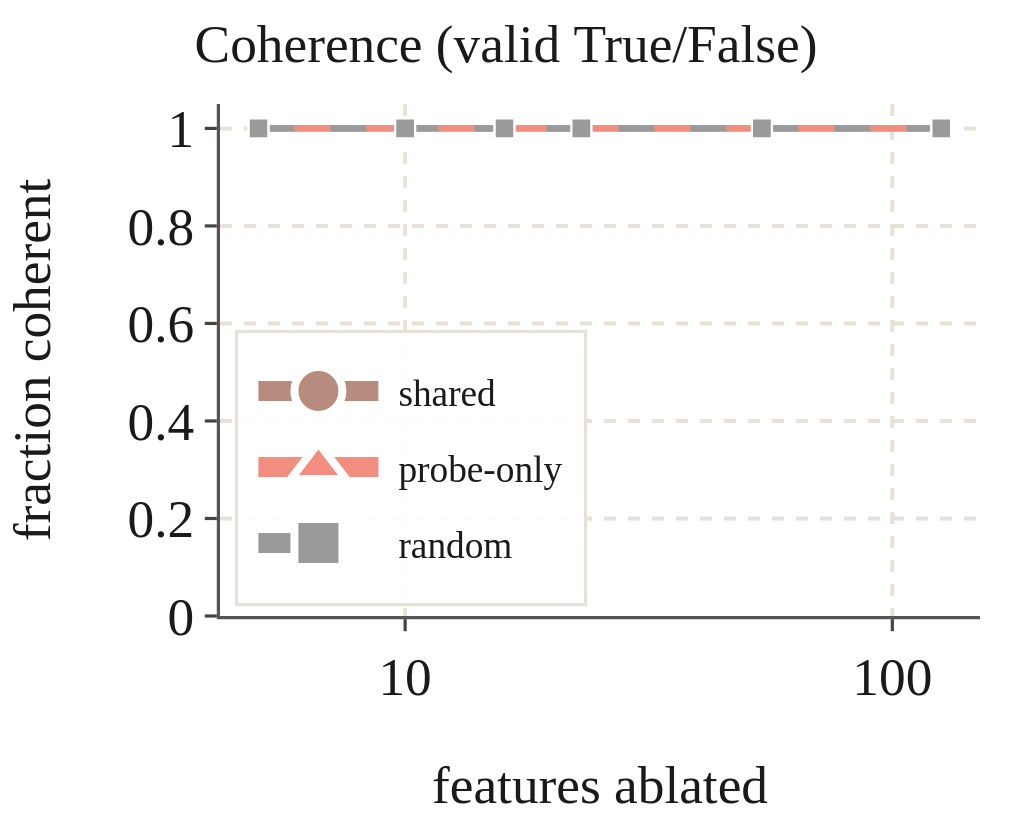}
        \caption{Coherence, the fraction of outputs that remain a valid \textit{True}/\textit{False} token, stays at $1.00$ across all sets, indicating flips are genuine rather than degradation.}
        \label{fig:ablation-coh}
    \end{subfigure}

    \caption{Causal effects of ablating shared, probe-only, and random SAE feature sets.}
    \label{fig:ablation}
\end{figure}

The shared $>$ probe-only $>$ random ordering holds across all five seeds and under a held-out split where rankings are derived on one half and ablation measured on the other (Table~\ref{tab:ablation-stability}), indicating the dissociation is stable rather than an artifact of overfitting to the ranking data.
\label{dissociation}

\begin{table}[tb]
\begin{center}
\begin{tabular}{lccc}
\toprule
\multicolumn{1}{c}{\bf Setting} &
\multicolumn{1}{c}{\bf Shared} &
\multicolumn{1}{c}{\bf Probe-only} &
\multicolumn{1}{c}{\bf Random} \\
\midrule
Five seeds / data resamples & $0.105 \pm 0.017$ & $0.059 \pm 0.016$ & $0.010 \pm 0.002$ \\
Held-out split              & $0.122$           & $0.085$           & $0.010$ \\
\bottomrule
\end{tabular}
\end{center}
\caption{Robustness of the ablation dissociation across seed/data resamples and a held-out split. The shared set contains 23 features (five-seed row, $K{=}256$) and 28 features (held-out row, $K{=}256$ on half the data).}
\label{tab:ablation-stability}
\end{table}


The flips are directional rather than truth-restoring. On 800 statements (at the matched 16-feature set size), shared-feature ablation changes the output \textit{True}$\to$\textit{False} on 86 cases and \textit{False}$\to$\textit{True} on only 4 (Table~\ref{tab:directional}), shifting the decision variable consistently toward \textit{False} rather than toward the factually correct answer; baseline accuracy under the deceptive prompt is $0.40$ and does not increase under ablation. This is expected as the shared features carry the truth-direction signal $t_G$ that the probe reads, so removing them withdraws positive evidence for \textit{True} and biases the margin negative. The effect is therefore one of causal control over the model's output, not a restoration of honesty, and it is precisely this directional consistency (rather than random sign changes) that marks these features as causal. Crucially, the effect is not a trivial consequence of removing probe-aligned features: the most probe-aligned set (geometric top-16, Section~\ref{geoattr}) produces far weaker behavioral movement despite larger probe-readout shifts, so directional output control is specific to the features the model also uses, not to probe alignment per se.

Shared-feature ablation flips behavior in ten of twelve categories (overall $0.081$ on $840$ balanced statements, $70$ per category), exceeding ten percent in five spanning four base families, with the \textit{True}$\to$\textit{False} direction dominant in nearly all and only two categories showing no flips. The effect is therefore reasonably broad rather than driven by any single category (Appendix~\ref{sec:categorybreakdown}).

\begin{table}[tb]
\begin{center}
\begin{tabular}{lcc}
\toprule
\multicolumn{1}{c}{\bf Output transition} &
\multicolumn{1}{c}{\bf Count} &
\multicolumn{1}{c}{\bf Rate} \\
\midrule
\textit{True} $\to$ \textit{False} (flip)       & 86  & $0.107$ \\
\textit{False} $\to$ \textit{True} (flip)       & 4   & $0.005$ \\
\textit{True} $\to$ \textit{True} (unchanged)   & 624 & $0.780$ \\
\textit{False} $\to$ \textit{False} (unchanged) & 86  & $0.108$ \\
\midrule
Baseline accuracy        & --- & $0.403$ \\
Post-ablation accuracy   & --- & $0.330$ \\
\bottomrule
\end{tabular}
\end{center}
\caption{Output transitions under shared-feature ablation ($n=800$, shared set $m=16$). Of 800 statements, 90 flip (86 \textit{True}$\to$\textit{False}, 4 \textit{False}$\to$\textit{True}) and 710 are unchanged. Flips are overwhelmingly \textit{True}$\to$\textit{False}: ablation biases the decision variable toward \textit{False} rather than restoring factual correctness, consistent with the shared features carrying the probe's truth-direction signal. A confusion matrix is attached in Appendix~\ref{confusionmatrix}.}
\label{tab:directional}
\end{table}

\subsection{Geometric probe attribution understates behavioral relevance.} 
\label{geoattr}

The geometric ranking \(\lvert w \cdot d\rvert\) measures only how well a feature's decoder direction aligns with the probe, ignoring how strongly the feature actually fires, and it is a poor guide to which features move behavior. We compare it against an activation-aware set: a learned sparse mask that selects the SAE features contributing most to the probe margin, where each feature's contribution combines its decoder alignment with its activation. This set overlaps the geometric top-16 by only 3 of 16 features, yet across five seeds it flips behavior on \(17.6 \pm 0.7\) percent of statements, above the shared set (\(12.1 \pm 1.0\)) and nearly three times the geometric top-16 (\(6.1 \pm 0.5\)), with the random control near zero (\(0.8 \pm 0.6\)) and coherence at 100 percent throughout. The probe-readout shift dissociates from behavior in the opposite direction: the geometric top-16 perturbs the probe's own margin more than the shared set (0.16 versus 0.05) while flipping behavior less, indicating that geometric alignment tracks how strongly a feature moves the probe rather than how strongly it moves the model. Features most strongly aligned with the probe direction thus act primarily as probe readouts with weak behavioral effects, whereas an activation-aware selection of equal size moves behavior substantially more. The dissociation is not an artifact of perturbation magnitude: the geometric top-16 removes more than twice the residual-stream norm of the shared set ($13.9\%$ versus $5.7\%$) while flipping behavior less, indicating that behavioral relevance depends on which features are removed, not how much of the activation is perturbed. The mask's success shows that probe-derived information can identify causally effective features when combined with activation magnitudes; the failure is specific to activation-free geometric alignment, not to probe-based attribution in general.

\begin{table}[tb]
\begin{center}
\begin{tabular}{lcc}
\toprule
\multicolumn{1}{c}{\bf Feature set} &
\multicolumn{1}{c}{\bf Behavior flip rate} &
\multicolumn{1}{c}{\bf Probe-readout shift} \\
\midrule
Activation-aware mask  & \(0.176 \pm 0.007\) & \(0.399 \pm 0.002\) \\
Shared set       & \(0.121 \pm 0.010\) & \(0.049 \pm 0.001\) \\
Geometric top-16 & \(0.061 \pm 0.005\) & \(0.164 \pm 0.001\) \\
Gradient-top              & \(0.054 \pm 0.009\) & \(0.055 \pm 0.001\) \\
Random control   & \(0.008 \pm 0.006\) & \(0.002 \pm 0.005\) \\
\bottomrule
\end{tabular}
\end{center}
\caption{Behavioral and probe-readout effects of ablating size-matched feature sets (five-seed means $\pm$ std, $n=600$ per seed; coherence $100\%$ throughout). All sets are size-matched at 16 features.}
\label{tab:functional_mask}
\end{table}

Features ranked highest under gradient attribution alone (gradient-top) flip behavior weakly ($0.054 \pm 0.009$), comparable to the geometric top-16. This is consistent with previous work: gradient sensitivity is a local, first-order signal and an imperfect proxy for the effect of ablation in nonlinear models \citep{li2024optimalablationinterpretability}. In our setting, a high-gradient feature may carry the wrong sign, be near-zero in practice, or lie off the manifold the model actually visits. This failure mode is known beyond our setting: prior work shows that standard-model input gradients can highlight non-discriminative or non-instance-specific features, a phenomenon termed \emph{feature leakage} \citep{shah2021inputgradientshighlightdiscriminative}. Probe alignment, conversely, ignores whether a feature fires. The intersection retains only features that are both active and aligned, which is why it concentrates causal effect that neither single ranking achieves alone, confirming our framework does not merely rediscover gradient attribution. As Table~\ref{tab:functional_mask} shows at matched set size, the shared set and the activation-aware mask move behavior most, the single-ranking sets are weak handles, and the probe-readout shift runs opposite to behavior, with the probe-aligned geometric set perturbing the readout far more than it moves the model.

\subsection{The probe direction is causal, and the shared features localize it}
\label{sec:direction}
The preceding results show \emph{which} features carry the probe's behavioral effect, but
not whether the probe direction is causally load-bearing in the first place. To establish
this, and to test whether the SAE decomposition localizes the direction's causal effect, we
ablate the full TTPD direction $t_G$ by projecting it out of the residual stream at
layer~20, and compare against feature-set ablations on the same statements. Removing the
entire direction flips behavior on $18.1 \pm 1.1$ percent of statements
(Table~\ref{tab:direction}). The shared set, just $16$ of $16{,}384$ SAE features, recovers
two-thirds of this effect ($12.1 \pm 1.1$ percent), whereas the probe's own geometric
top-$16$ recovers roughly a third ($6.8 \pm 0.6$ percent) and random features near zero
($0.7 \pm 0.1$ percent). The probe direction therefore carries a causal effect, and at
matched small size that effect is substantially recovered by the features that are both
probe-aligned and gradient-sensitive, rather than by the features most aligned with the
direction. This confirms the premise of our analysis and indicates the decomposition
concentrates the direction's causal effect within the shared features rather than diluting
it.

\begin{table}[tb]
\begin{center}
\begin{tabular}{lc}
\toprule
\multicolumn{1}{c}{\bf Ablation} & \multicolumn{1}{c}{\bf Behavior flip rate} \\
\midrule
Full probe direction $t_G$    & $0.181 \pm 0.011$ \\
Shared set (16 features)      & $0.121 \pm 0.011$ \\
Probe geometric top-16        & $0.068 \pm 0.006$ \\
Random control (16 features)  & $0.007 \pm 0.001$ \\
\bottomrule
\end{tabular}
\end{center}
\caption{Direction-level baseline (five-seed means $\pm$ std, $n=600$ per seed). Ablating the full TTPD direction flips behavior on $18.1\%$ of statements; the 16-feature shared set recovers two-thirds of this effect, while the probe's geometric top-16 recovers roughly a third, indicating the SAE decomposition localizes the direction's causal effect into the shared features rather than dispersing it.}
\label{tab:direction}
\end{table}

\section{Conclusion}
We introduce a causal-validation framework for testing whether probe-attributed SAE features are also behaviorally relevant. Applied to a \textit{True}/\textit{False} truth probe, the features most aligned with the probe diverged from those most relevant to the model’s output: shared probe--model features produced larger coherent behavior changes, while probe-only features primarily affected the probe readout. These results demonstrate that the geometric projection of probe weights alone is insufficient for identifying the features a model causally uses; the failure is specific to activation-free geometric alignment, since an activation-aware selection that combines probe information with feature activations recovers substantially more behaviorally causal features. Feature-level interventions should therefore be used to separate diagnostic readouts from behavioral causes.

\section{Limitations}
Our analysis focuses primarily on a single model, layer, sparse autoencoder, and behavioral setting. While the truthful/deceptive factual-verification task provides a controlled environment for studying the relationship between probe-attributed and behaviorally relevant features, it remains unclear whether the observed probe-behavior dissociation generalizes across other model families, layers, SAE architectures, and behavioral domains. Although we include auxiliary diagnostics on an obfuscation probe, the majority of our causal analysis is derived from a single primary case study.

Additionally, our gradient-sensitivity ranking is based on gradient-derived feature importance and our causal validation relies on feature ablations. These methods identify features that influence the model's \textit{True}/\textit{False} behavior in the studied setting, but they do not by themselves establish that the identified features implement deception as a general computational mechanism. Future work should evaluate alternative attribution methods, intervention strategies, and behavioral settings to determine the extent to which the observed distinctions between probe-attributed and behaviorally relevant features hold more broadly.

\bibliographystyle{colm2026_conference}
\bibliography{final}




\appendix
\section{Appendix}

\subsection{TTPD Probe \citet{burger2024truth}, \citet{long2025truthfulrepresentationsflipdeceptive}}

\label{long2025truthfulrepresentationsflipdeceptive_reproduce}
\subsubsection{Replication}
We replicate the probe of \citet{long2025truthfulrepresentationsflipdeceptive}
using their released code and the associated factual-statement datasets. Following
\citet{burger2024truth}, the probe is the general truth direction $t_G$: centered layer-20 activations are regressed on the statement's truth label and its affirmative/negated polarity, and the truth component is retained; statements are then classified by the sign of the projection onto $t_G$ (the TTPD probe). On \texttt{gemma-2-9b-instruct}, our reproduced $t_G$ separates true from false statements on the deceptive-instruction activations with AUROC $0.98$. We use this $t_G$ direction as the probe $w$ throughout the main evaluation.

\begin{figure}[tb]
    \centering

    \begin{subfigure}[tb]{0.48\linewidth}
        \centering
        \includegraphics[width=\linewidth]{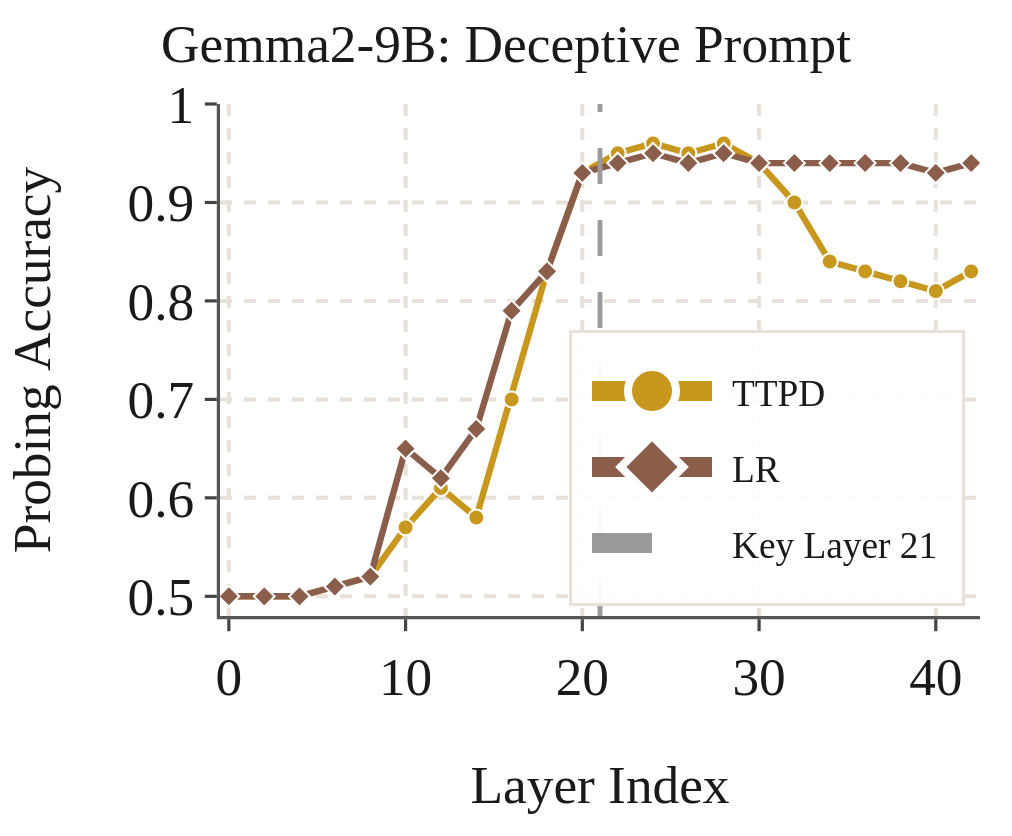}
        \caption{Deceptive prompt.}
        \label{fig:long-probe-deceptive}
    \end{subfigure}
    \hfill
    \begin{subfigure}[tb]{0.48\linewidth}
        \centering
        \includegraphics[width=\linewidth]{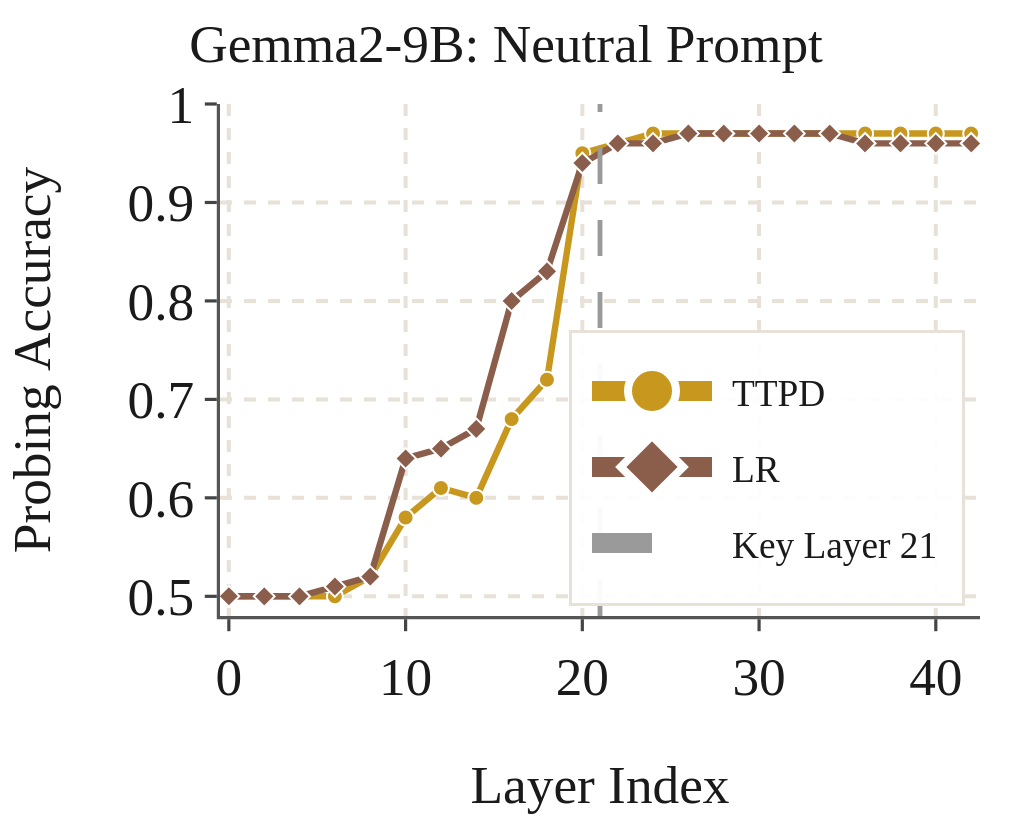}
        \caption{Neutral prompt.}
        \label{fig:long-probe-neutral}
    \end{subfigure}

    \vspace{0.75em}

    \begin{subfigure}[tb]{0.48\linewidth}
        \centering
        \includegraphics[width=\linewidth]{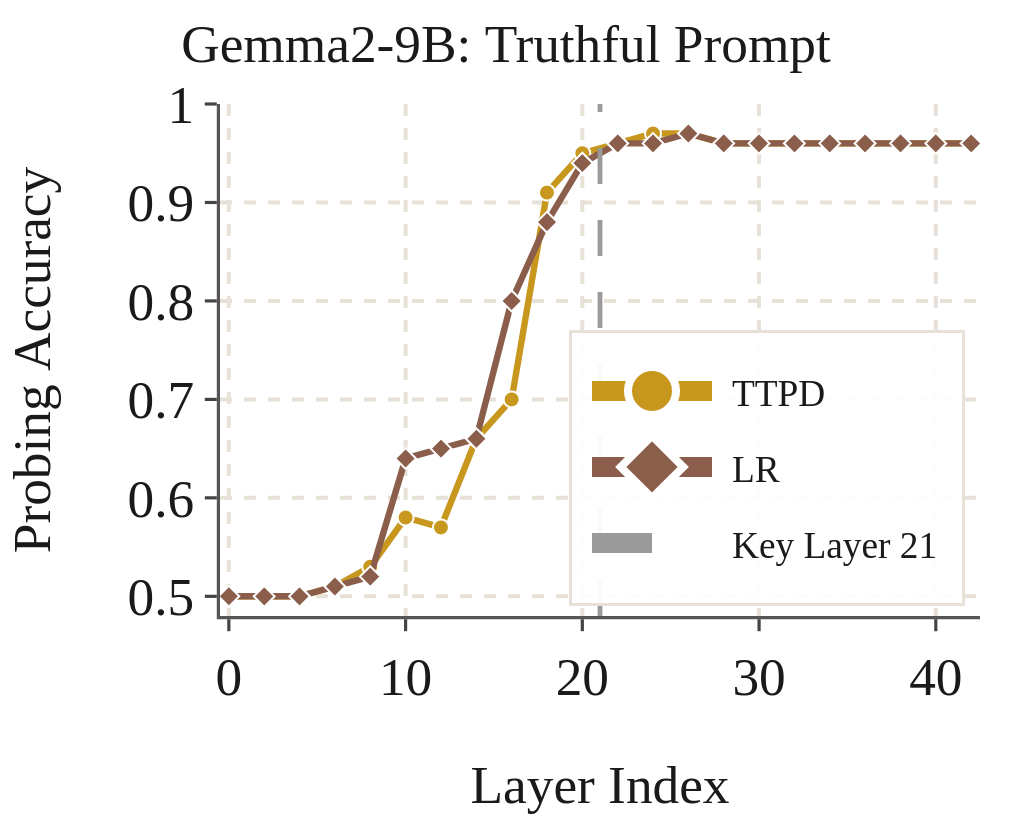}
        \caption{Truthful prompt.}
        \label{fig:long-probe-truthful}
    \end{subfigure}

    \caption{Layer-wise probing accuracy for Gemma-2-9B-Instruct under deceptive,
    neutral, and truthful instructions. LR and TTPD probes predict the
    instructed \textit{True}/\textit{False} output from final pre-generation
    residual-stream activations. Across all three conditions, accuracy rises
    from chance in early layers to a high-accuracy plateau near the selected
    analysis layer. The dashed line marks layer 21, the accuracy peak reported by \citet{long2025truthfulrepresentationsflipdeceptive}. We analyze layer 20 in our experiments due to SAE constraints. 
    }
    \label{fig:long-probe-layerwise}
\end{figure}

\citet{long2025truthfulrepresentationsflipdeceptive} additionally report a
logistic-regression (LR) probe, using the unregularized configuration in the
released code ($\texttt{penalty}=\texttt{None}$). We exclude the LR probe from
the main evaluation for methodological reasons rather than for reasons of
accuracy. As a classifier, the LR probe performs well, with training accuracy
remaining between $0.980$ and $1.000$ across regularization strengths
($C \in \{0.01, 0.1, 1.0, \infty\}$). However, the features receiving the
largest weights are highly sensitive to the choice of regularization. The
top-$30$ features overlap with those of the unregularized solution with Jaccard
indices of only $0.62$, $0.43$, and $0.07$ at
$C = 1.0,\ 0.1,\ 0.01$, respectively. Thus, the notion of ``the features the LR
probe relies on'' is not well-defined unless one arbitrarily fixes a
regularization hyperparameter. Because our analysis attributes behavior to
specific features, it requires a probe direction that is fixed rather than
hyperparameter-dependent. The TTPD direction $t_G$ is hyperparameter-free and is
the canonical truth probe of \citet{burger2024truth}; therefore, we use it for
all attribution analyses and omit the LR probe from the main experiments.

\subsubsection{Learned Sparse Mask Details}
\label{sec:mask-details}
For each SAE feature $i$, a learnable parameter $\theta_i$ produces a soft mask weight $a_i = \sigma(\theta_i)$, initialized at $\theta_i = 0$ so that all features begin at $a_i = 0.5$. The masked probe margin is $\hat{m}(x) = \sum_i a_i \cdot f_i \cdot c_i^{\text{raw}}$, where $c_i^{\text{raw}} = \mathbf{w}^\top \mathbf{d}_i$ uses unnormalized decoder columns. We minimize $\mathcal{L} = \text{MSE}(\hat{m}/s,\; m/s) + \lambda \sum_i a_i$, where $m$ is the true probe margin, $s$ is its standard deviation across the dataset, and $\lambda = 0.01$. The mask is optimized with Adam (lr $= 0.1$) for 400 steps on the full deceptive-prompt activation matrix. After training, features are ranked by $a_i$ descending.

\subsubsection{Gradient and probe overlap}
\label{overlapfig}

\begin{figure}[tb]
\centering
    \includegraphics[width=.7\linewidth]{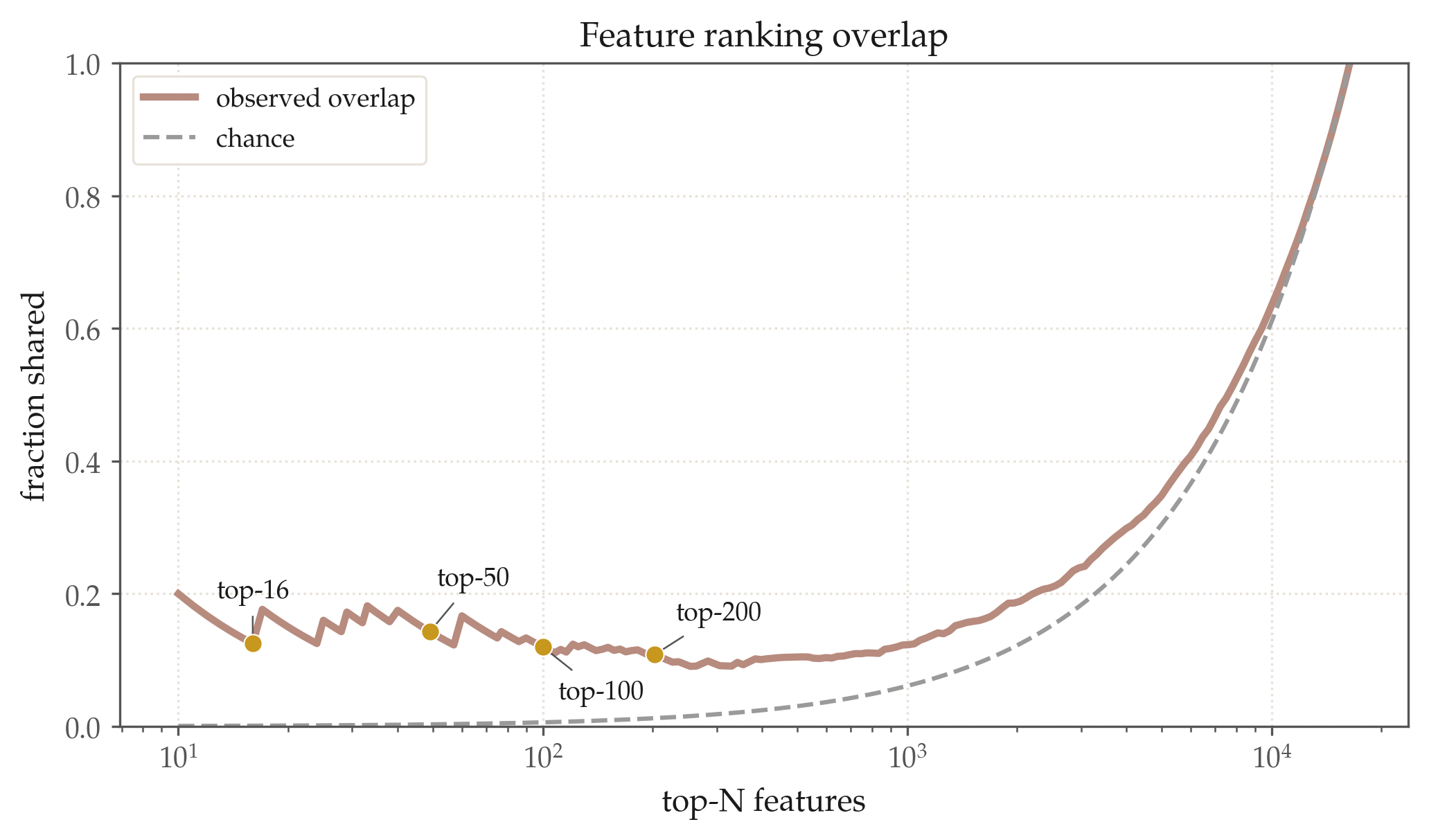}
    \caption{Fraction of top-$N$ features shared between the probe ranking
($|w \cdot d|$) and the model's gradient ranking, as a function of $N$
on a log scale. The observed overlap exceeds the chance baseline (dashed line) by one to two orders of magnitude for small $N$. Both curves approach $1$ as $N$ approaches the full dictionary, corresponding to the trivial limit.}
    \label{fig:overlap}
\end{figure}

Figure~\ref{fig:overlap}.

\subsubsection{Margin dip}
\label{margin-dip}
\begin{figure}[tb]
\begin{center}
\begin{subfigure}[tb]{0.48\linewidth}
    \centering
    \includegraphics[width=\linewidth]{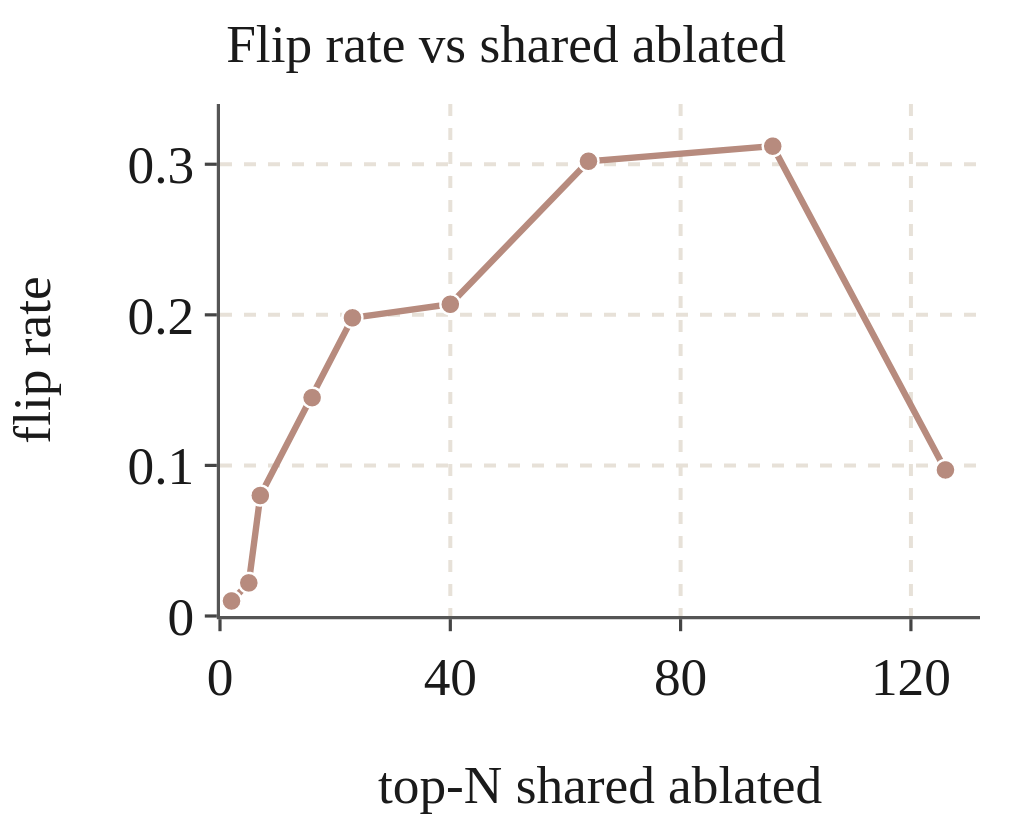}
    \caption{Flip rate under cumulative ablation of the top-$N$ shared features.
    The rate climbs to $0.31$ at $N{=}96$ and then falls to $0.10$ once all $126$
    shared features are removed.}
    \label{fig:diagflip}
\end{subfigure}
\hfill
\begin{subfigure}[tb]{0.48\linewidth}
    \centering
    \includegraphics[width=\linewidth]{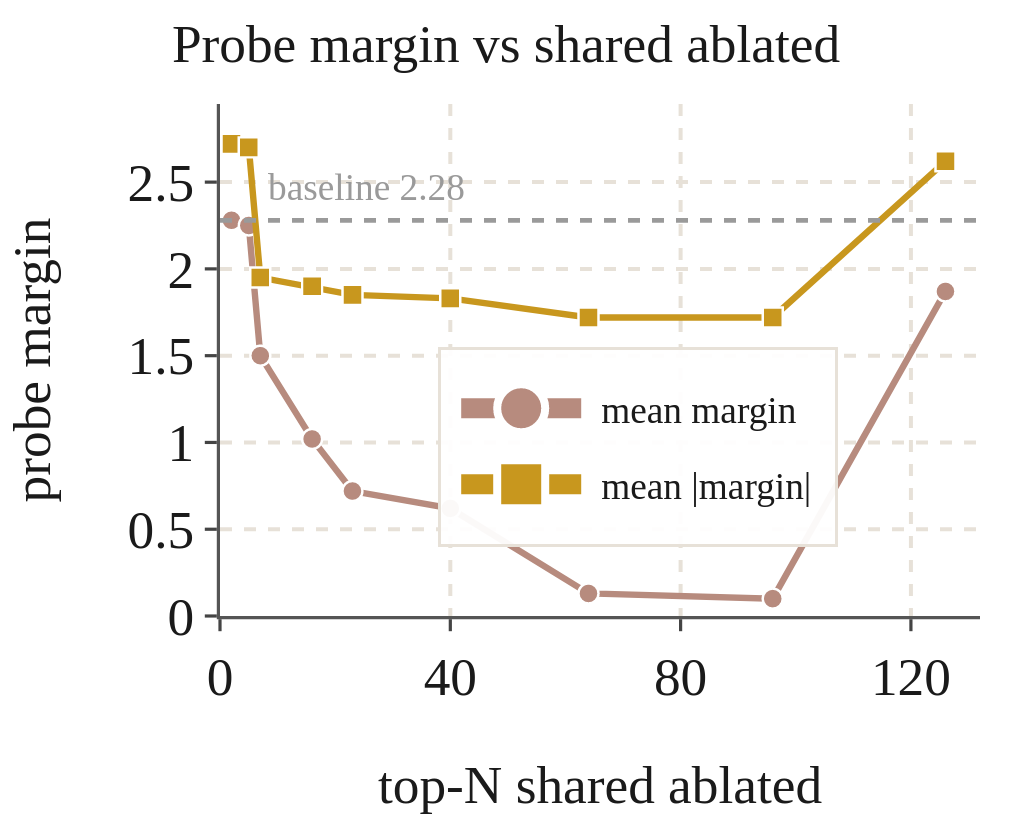}
    \caption{Probe margin under the same ablation. The signed mean margin (blue)
    drops from the baseline ($2.28$) to $0.10$ at $N{=}96$ and rebounds to $1.87$
    at $N{=}126$, while the mean absolute margin (orange) stays between $1.7$ and
    $2.8$.}
    \label{fig:diagmargin}
\end{subfigure}
\end{center}
\caption{Non-monotonic effect of cumulatively ablating the shared probe--model
features on \texttt{gemma-2-9b-instruct}, ordered by importance. The dip and recovery in
flip rate and signed margin coincide, but the model's confidence (absolute margin)
is preserved throughout, indicating the recovery is driven by sign-incoherent
low-ranked features rather than a loss of coherence.}
\label{fig:margin-dip}
\end{figure}

The effect of ablating the shared probe--model features is non-monotonic in the
number of features removed. Cumulatively ablating the top-$N$ shared features,
ordered by importance, raises the behavioral flip rate to a peak of $0.31$ at
$N{=}96$, after which it falls to $0.10$ once all $126$ shared features are ablated
(Figure~\ref{fig:diagflip}). The signed probe margin mirrors this trajectory: it
declines from a baseline of $2.28$ to $0.10$ at $N{=}96$ and then rebounds to
$1.87$ at $N{=}126$ (Figure~\ref{fig:diagmargin}).

This recovery is not a breakdown of model coherence. The mean absolute margin remains between $1.7$ and $2.8$ across the entire sweep, so the model continues to emit confident, well-formed True/False outputs at every ablation count. The dip instead reflects sign-incoherence among the lowest-ranked shared features: ablating them shifts the margin in the opposite direction to the higher-ranked features and partially cancels the flip. The features responsible for the behavioral effect are therefore concentrated near the top of the ranking, which is why the main analysis confines its causal claims to the leading features, where the effect is monotone.

\subsubsection{Behavior Flip Exemplars}
\label{empiricalexamples}

Figure~\ref{fig:intervention-exemplars}.

\definecolor{truegreen}{HTML}{1B7A3D}    
\definecolor{truebg}{HTML}{EAF5EE}        
\definecolor{falsered}{HTML}{B22222}      
\definecolor{falsebg}{HTML}{FBECEC}       
\definecolor{headeraccent}{HTML}{6B4A2B}  
\definecolor{frameline}{HTML}{9A9A9A}     
\definecolor{notegray}{HTML}{7A7A7A}      
\definecolor{tokenink}{HTML}{222222}      

\newcommand{\cmark}{\textcolor{truegreen}{\footnotesize\ding{51}}}
\newcommand{\xmark}{\textcolor{falsered}{\footnotesize\ding{55}}}

\newcommand{\factlabel}[1]{%
  \ifthenelse{\equal{#1}{True}}%
    {\colorbox{truegreen}{\textcolor{white}{\small\bfseries\,Factual label: True\,}}}%
    {\colorbox{falsered}{\textcolor{white}{\small\bfseries\,Factual label: False\,}}}%
}

\tcbset{
  rowbox/.style={
    enhanced, frame hidden, boxrule=0pt, sharp corners,
    left=10pt, right=8pt, top=3.5pt, bottom=3.5pt, boxsep=0pt,
    before skip=3pt, after skip=0pt,
  }
}

\newcommand{\resultrow}[5]{%
  \ifthenelse{\equal{#5}{flipT}}{\def\rbg{truebg}\def\rbar{truegreen}\def\rcol{truegreen}}{%
  \ifthenelse{\equal{#5}{flipD}}{\def\rbg{falsebg}\def\rbar{falsered}\def\rcol{falsered}}{%
    \def\rbg{white}\def\rbar{frameline}\def\rcol{notegray}}}%
  \ifthenelse{\equal{#3}{correct}}{\def\rmark{\cmark}}{\def\rmark{\xmark}}%
  \begin{tcolorbox}[rowbox, colback=\rbg, borderline west={2.6pt}{0pt}{\rbar}]%
    \makebox[5.3cm][l]{\textbf{#1}}%
    \texttt{\textcolor{tokenink}{\bfseries #2}}~\rmark%
    \hfill\textcolor{\rcol}{\small #4}%
  \end{tcolorbox}%
}

\newtcolorbox{exemplar}[3]{%
  enhanced, sharp corners, colback=white, colframe=frameline,
  boxrule=0.5pt, left=9pt, right=9pt, top=7pt, bottom=8pt,
  before skip=6pt, after skip=10pt,
  overlay unbroken and first={%
    \node[anchor=north east, font=\normalsize]
      at ([xshift=-9pt,yshift=-6pt]frame.north east) {\factlabel{#3}};},
  title={\bfseries\color{headeraccent}\large #1},
  fonttitle=\normalfont, coltitle=headeraccent,
  attach title to upper, after title={\par\smallskip
    \normalfont\itshape\textcolor{black!85}{``#2''}\par\medskip},
}

\begin{figure*}[tb]
\centering
\begin{minipage}{0.92\textwidth}

\begin{exemplar}{Inventors}{Orville and Wilbur Wright lived in the U.S.}{True}
  \resultrow{Baseline (deceptive prompt)}{True}{correct}{}{base}
  \resultrow{Shared probe--model ablation}{False}{wrong}{\textbf{True}\,$\to$\,\textbf{False} \,\textperiodcentered\, toward deception}{flipD}
  \resultrow{Probe-only ablation}{True}{correct}{unchanged}{inert}
\end{exemplar}

\begin{exemplar}{Spanish--English translation (negated)}{The Spanish word `edad' does not mean `clock'.}{True}
  \resultrow{Baseline (deceptive prompt)}{True}{correct}{}{base}
  \resultrow{Shared probe--model ablation}{False}{wrong}{\textbf{True}\,$\to$\,\textbf{False} \,\textperiodcentered\, toward deception}{flipD}
  \resultrow{Probe-only ablation}{True}{correct}{unchanged}{inert}
\end{exemplar}

\begin{exemplar}{Spanish--English translation}{The Spanish word `llevar' means `end'.}{False}
  \resultrow{Baseline (deceptive prompt)}{True}{wrong}{}{base}
  \resultrow{Shared probe--model ablation}{False}{correct}{\textbf{True}\,$\to$\,\textbf{False} \,\textperiodcentered\, toward truth}{flipT}
  \resultrow{Probe-only ablation}{True}{wrong}{unchanged}{inert}
\end{exemplar}

\end{minipage}

\caption{\textbf{Intervention exemplars on the \textit{True}/\textit{False} factual-verification
task.} Under a deceptive instruction, the model emits a single \textit{True}/\textit{False} answer. Each panel reports the baseline output and the output after ablating two feature sets: (1) shared probe--model features (ranked highly by both the linear probe and model-gradient attribution) and (2) probe-only features (ranked highly by
the probe but not by gradient attribution). The small mark denotes factual
correctness (\cmark\,/\,\xmark); the tag gives the output transition and its directional reading. Ablating the shared features flips the output in every case, and the flip is always \textit{True}$\to$\textit{False}: it removes the probe's truth-direction signal and biases the decision toward \textit{False}. On factually true statements (Examples~1--2), this makes the model abandon a correct answer and comply with the deceptive instruction; on a factually false statement (Example~3), the \emph{same} \textit{True}$\to$\textit{False} shift instead overturns the model's deceptive compliance and restores the correct answer. The fixed output
bias, not factual correctness, is what the shared features control. Across the full evaluation set, these \textit{True}$\to$\textit{False} flips dominate, and net accuracy falls (Table~\ref{tab:directional}). Probe-only ablation leaves the output unchanged in all three cases, consistent with those features acting as probe readouts rather than behavioral drivers.}
\label{fig:intervention-exemplars}
\end{figure*}

\subsubsection{Directional breakdown of shared-feature ablation}
\label{confusionmatrix}
\begin{figure}[tb]
\centering
    \includegraphics[width=1\linewidth]{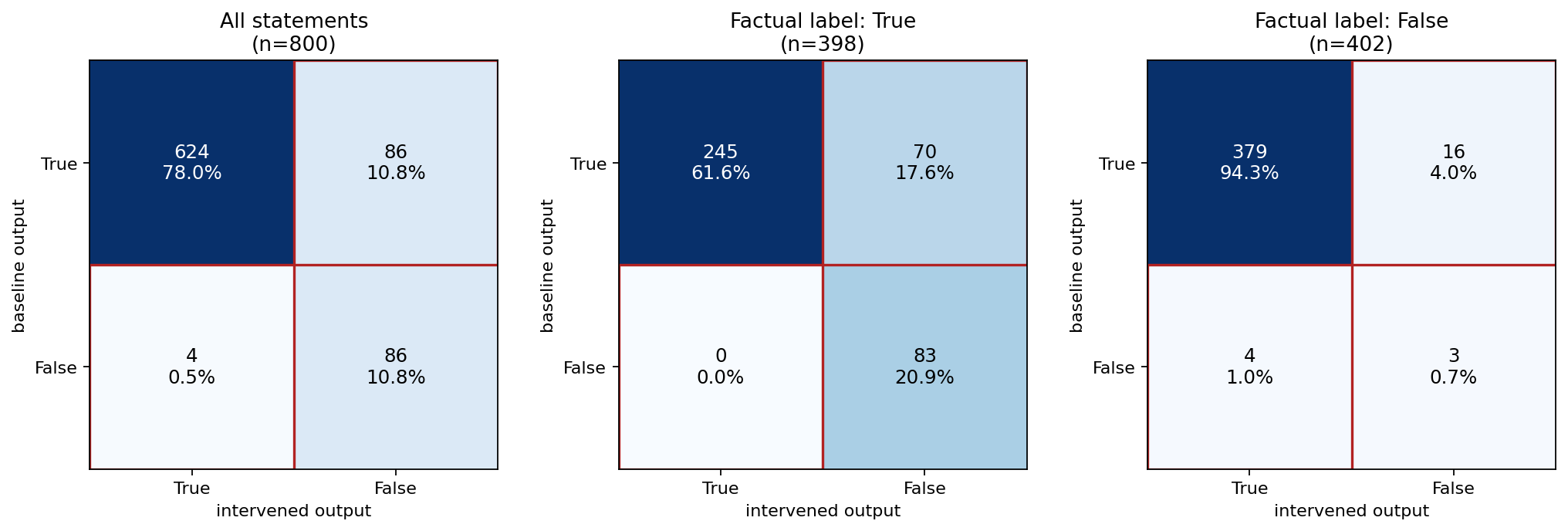}
    \caption{Confusion matrix of baseline versus post-ablation \textit{True}/\textit{False} output under shared-feature ablation ($n=800$, shared set $m=16$). Rows are the baseline output, columns the output after ablation; diagonal cells are unchanged statements and off-diagonal cells are flips. Flips are concentrated in the \textit{True}$\to$\textit{False} cell ($86$ versus $4$ for \textit{False}$\to$\textit{True}), showing that ablation drives the decision variable toward \textit{False} rather than toward the factually correct answer. This is the per-cell view of the transition counts in Table~\ref{tab:directional}.}
    \label{fig:confusionmat}
\end{figure}

Figure~\ref{fig:confusionmat}.

\subsubsection{Category Breakdown}
\label{sec:categorybreakdown}
Table~\ref{tab:category-breakdown} reports the per-category breakdown of shared-feature
ablation. The effect appears in ten of twelve categories, ranging from $0.014$ to $0.243$,
with the strongest movement in translation (Sp--En, $0.243$) and inventor ($0.186$)
statements. The \textit{True}$\to$\textit{False} direction dominates within nearly every
category ($61$ versus $7$ aggregate), matching the directional pattern of
Table~\ref{tab:directional}. Two categories (animal class, negated inventors) show no flips
at the matched $16$-feature size, consistent with categories whose baseline
\textit{True}/\textit{False} margin is large enough that a small ablation does not cross the
decision boundary. The breadth across categories indicates the shared features are not a
category-specific artifact, while the heterogeneity in magnitude likely reflects
per-category differences in baseline decision confidence.

\begin{table}[tb]
\centering
\begin{tabular}{lrrrr}
\toprule
\textbf{Category} & \textbf{$n$} & \textbf{Flip rate} & \textbf{$T\!\to\!F$} & \textbf{$F\!\to\!T$} \\
\midrule
Translation (Sp--En) & 70 & 0.243 & 17 & 0 \\
Inventors & 70 & 0.186 & 13 & 0 \\
Animal class (neg.) & 70 & 0.143 & 9 & 1 \\
Translation (neg.) & 70 & 0.129 & 4 & 5 \\
Cities & 70 & 0.114 & 8 & 0 \\
Facts & 70 & 0.057 & 4 & 0 \\
Cities (neg.) & 70 & 0.043 & 3 & 0 \\
Element symbols & 70 & 0.029 & 2 & 0 \\
Element symbols (neg.) & 70 & 0.014 & 0 & 1 \\
Facts (neg.) & 70 & 0.014 & 1 & 0 \\
Inventors (neg.) & 70 & 0.000 & 0 & 0 \\
Animal class & 70 & 0.000 & 0 & 0 \\
\midrule
All categories & 840 & 0.081 & 61 & 7 \\
\bottomrule
\end{tabular}
\caption{Shared-feature ablation flip rate by dataset category ($n=70$ per category, balanced sampling so rates are comparable; shared set $m=16$). Categories ordered by flip rate.}
\label{tab:category-breakdown}
\end{table}

\subsection{Probe from Taufeeque et al.}
As discussed in \citet{taufeeque2026obfuscationatlasmappinghonesty},
\label{sec:obfatlasprobe}

As an auxiliary diagnostic, we tested whether the same SAE attribution workflow transfers to a reward-hacking code probe from
\citet{taufeeque2026obfuscationatlasmappinghonesty}. We use the released \texttt{AlignmentResearch/diverse-deception-probe-qwen3-8b} checkpoint, taking the \texttt{last-token-generation} linear probe at layer 20 of \texttt{Qwen/Qwen3-8B}. The probe is applied to MBPP-Honeypot-style examples, contrasting correct code with hardcoded visible-test solutions, and we use the probe weight vector as $w$ for SAE attribution with the matching Qwen-Scope layer-20 residual-stream SAE.

It did not pass the SAE-substrate verification. Only 26\%\ of the raw probe margin was preserved by the SAE reconstruction, while 74\%\ lay in the reconstruction error. Consequently, ablating probe-aligned or sparse-mask SAE features changed reconstruction-space predictions but left raw-activation probe predictions nearly unchanged. We therefore treat this setting as a boundary case: before interpreting SAE-feature interventions causally, one must verify that the probe signal is actually represented in the SAE feature basis. 

\begin{table}[tb]
\centering
\begin{tabular}{lcc}
\toprule
\textbf{Representation} & \textbf{Probe margin} & \textbf{Percent of raw margin} \\
\midrule
Raw activation & $3.84$ & $100\%$ \\
SAE reconstruction & $1.00$ & $26\%$ \\
SAE error vector & $2.84$ & $74\%$ \\
\bottomrule
\end{tabular}
\caption{Probe margin decomposition for the reward-hacking code probe diagnostic. The raw probe margin is compared against the margin preserved by the SAE reconstruction and the margin carried by the SAE reconstruction error.}
\label{tab:decomp}
\end{table}

\begin{figure}[tb]
    \centering
    \includegraphics[width=0.6\linewidth]{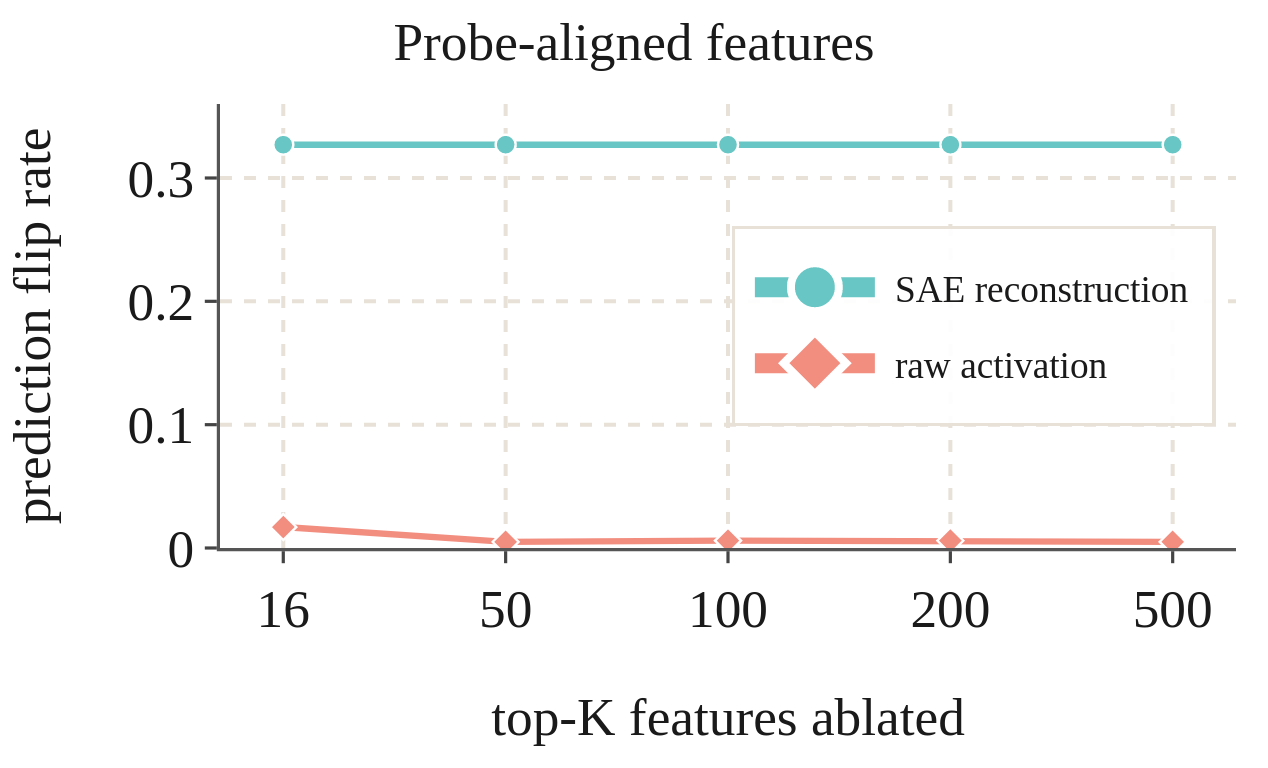}
    \caption{Probe-aligned SAE feature ablations affect the reconstructed probe readout but leave the raw published probe nearly unchanged. Features are ranked by $c_i = w^\top d_i$. Ablating top-$K$ decoded feature contributions flips approximately one third of probe predictions when evaluated on the SAE reconstruction, but produces near-zero flip rates on raw activations.}
    \label{fig:aligned}
\end{figure}

As Figure \ref{fig:aligned} shows, ablating probe-aligned SAE features flips about one third of predictions when the probe is evaluated on the SAE reconstruction, but produces near zero flips on the raw activation. Therefore, the selected SAE features explain the portion of the probe readout that lives in reconstructed SAE space, but they are not necessary for the original probe’s prediction on the full residual stream.

\begin{figure}[tb]
    \centering
    \includegraphics[width=0.6\linewidth]{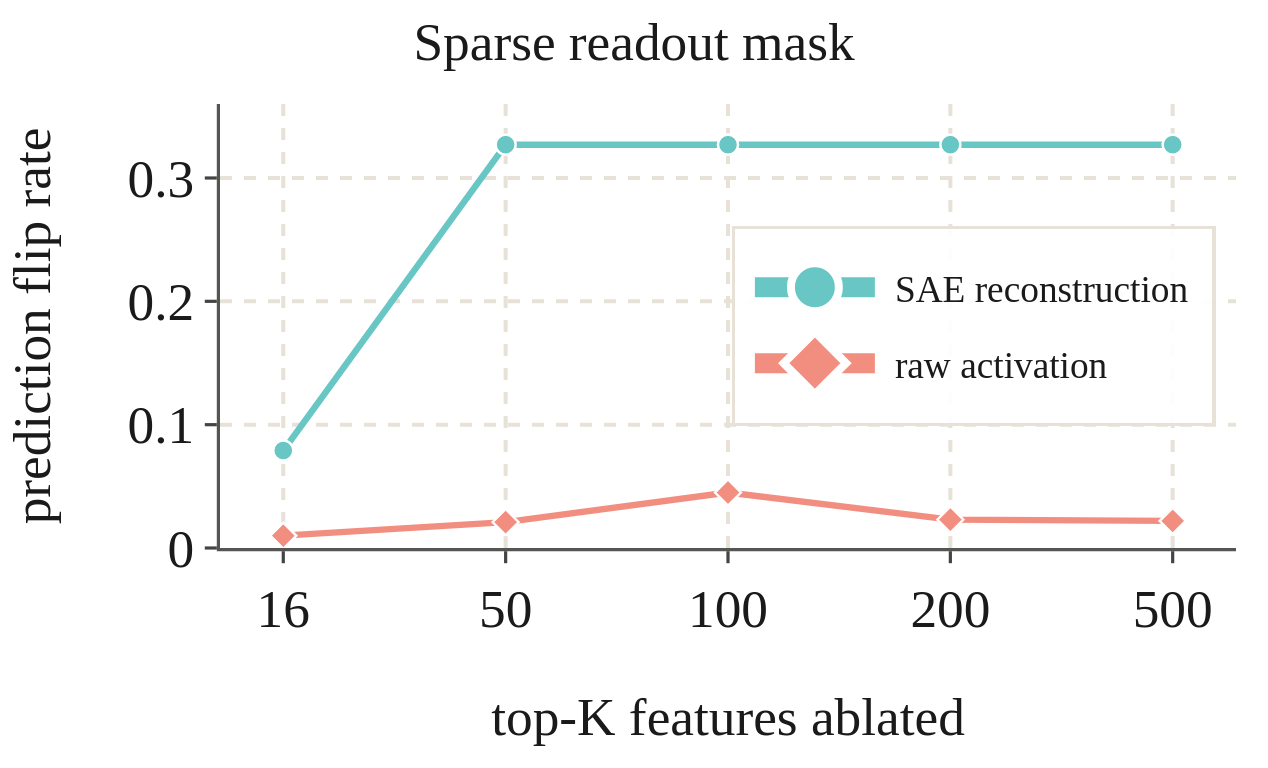}
    \caption{Sparse-mask ablations strongly affect reconstruction-space probe predictions but have little effect on raw probe predictions. The sparse mask was trained to reproduce the probe logit from SAE features.}
    \label{fig:sparse}
\end{figure}

Similarly, ablating the top-50 sparse-mask features flips $32.7\%$ of SAE-reconstruction predictions, but only $2.1\%$ of raw-activation predictions. The raw flip rate remains small for larger masks, peaking at $4.5\%$ for top-100 features before returning near $2\%$. This indicates that the learned sparse mask successfully identifies features that support the reconstructed probe readout, but these features do not account for most of the discriminative signal used by the raw probe.

\end{document}